\documentclass[sigconf]{acmart}

\usepackage{booktabs} % For formal tables
\usepackage{bm}
\usepackage{multirow}
\usepackage{CJKutf8}

\setcopyright{rightsretained}
\copyrightyear{2019}
\acmYear{2019} 
\setcopyright{iw3c2w3}
\acmConference[WWW '19]{Proceedings of the 2019 World Wide Web Conference}{May 13--17, 2019}{San Francisco, CA, USA}
\acmBooktitle{Proceedings of the 2019 World Wide Web Conference (WWW '19), May 13--17, 2019, San Francisco, CA, USA}
\acmPrice{}
\acmDOI{10.1145/3308558.3313570}
\acmISBN{978-1-4503-6674-8/19/05}

\begin{document}
\title{Domain-Constrained Advertising Keyword Generation}

\author{Hao Zhou$^{1}$, Minlie Huang$^{1,*}$, Yishun Mao$^{2}$, Changlei Zhu$^{2}$, Peng Shu$^{2}$, Xiaoyan Zhu$^{1}$}

\affiliation{%
  \department{$^{1}$Institute for Artificial Intelligence, 
State Key Lab of Intelligent Technology and Systems}
  \department{Beijing National Research Center for Information Science and Technology}
  \department{Department of Computer Science and Technology, Tsinghua University, Beijing 100084, China}
  \institution{$^{2}$Sogou Inc., Beijing, China}
}
\email{tuxchow@gmail.com;aihuang@tsinghua.edu.cn;ps-adwr@sogou-inc.com;}
\email{zhuchanglei@sogou-inc.com;shupeng203672@sogou-inc.com	;zxy-dcs@tsinghua.edu.cn}

\titlenote{Corresponding author: Minlie Huang, aihuang@tsinghua.edu.cn.}

\renewcommand{\shortauthors}{Zhou et al.}

\begin{abstract}
Advertising (ad for short) keyword suggestion is important for sponsored search to improve online advertising and increase search revenue. There are two common challenges in this task. First, {\bf the keyword bidding problem}: hot ad keywords are very expensive for most of the advertisers because more advertisers are bidding on more popular keywords, while unpopular keywords are difficult to discover. As a result, most ads have few chances to be presented to the users. Second, {\bf the inefficient ad impression issue}: a large proportion of search queries, which are unpopular yet relevant to many ad keywords, have no ads presented on their search result pages. Existing retrieval-based or matching-based methods either deteriorate the bidding competition or are unable to suggest novel keywords to cover more queries, which leads to inefficient ad impressions. 

To address the above issues, this work investigates to use generative neural networks for keyword generation in sponsored search. Given a purchased keyword (a word sequence) as input, our model can generate a set of keywords that are not only relevant to the input but also satisfy the domain constraint which enforces that the domain category of a generated keyword is as expected. 
Furthermore, a reinforcement learning algorithm is proposed to adaptively utilize domain-specific information in keyword generation. Offline evaluation shows that the proposed model can generate keywords that are diverse, novel, relevant to the source keyword, and accordant with the domain constraint. Online evaluation shows that generative models can improve coverage (COV), click-through rate (CTR), and revenue per mille (RPM) substantially in sponsored search.

\end{abstract}




\maketitle

\section{Introduction}

\begin{figure}[!htp]
  \centering
  \includegraphics[width=0.95\linewidth]{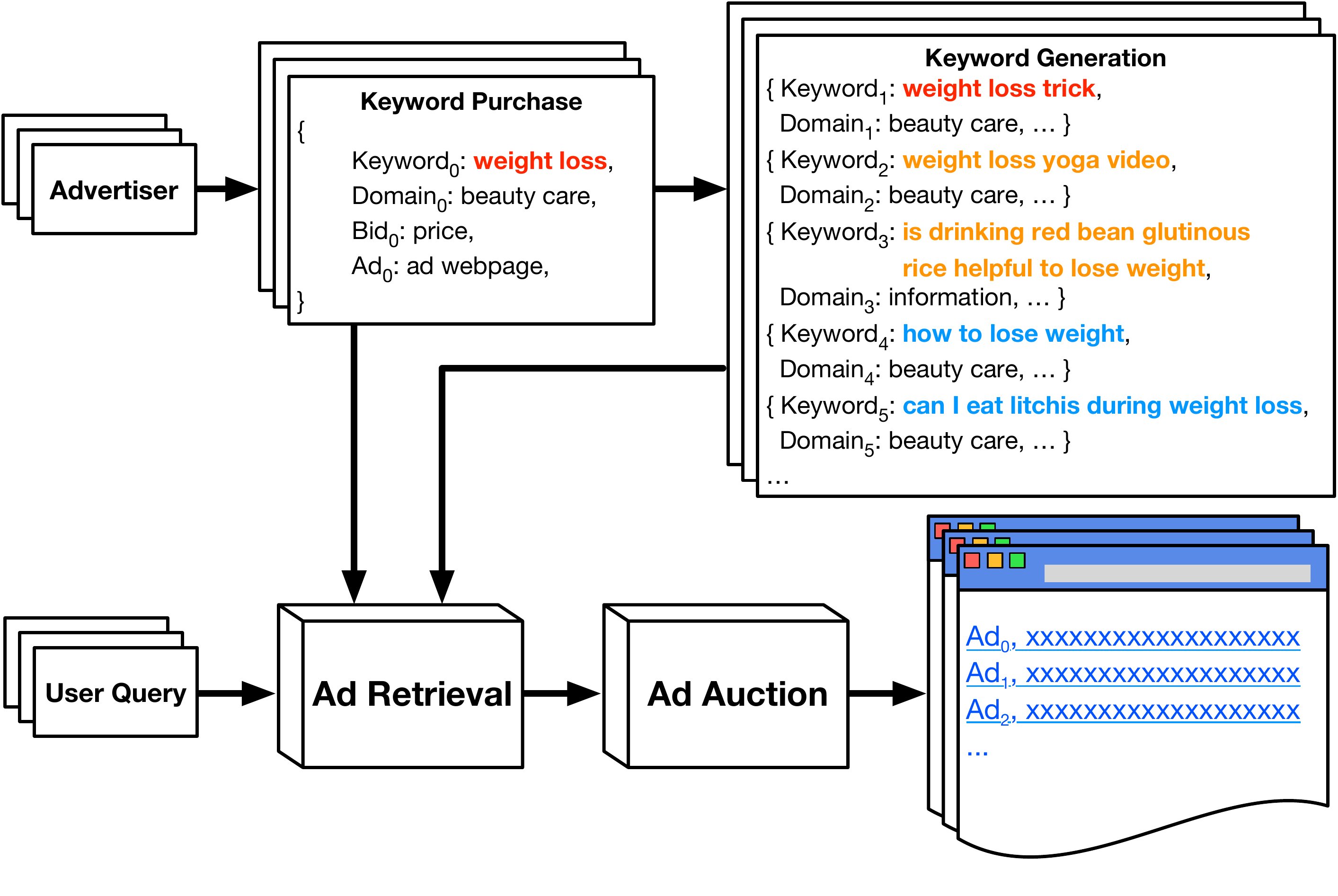}% 1\linewidth
  \caption{The sponsored search with keywords generated by our model. The red/orange/blue keywords are the keywords which have more than 3/less than 3/no ads on the impression webpage respectively. 
  }
  \label{fig:process}
\end{figure}

Advertising (ad for short) keyword suggestion is an important task for sponsored search which is one of the major types of online advertising and the major source of revenue for search companies. In sponsored search, a search engine first retrieves a set of advertisements whose keywords match a user issued query. It then ranks these advertising candidates according to an auction process by considering both the ad quality and the bid price of each ad~\cite{aggarwal2006truthful}. Finally, the chosen advertisement is presented in a search result page. 
Therefore, ad keywords are vital for ads to gain access to the auction process and have the chance to be displayed on a search result page. 

However, there are two common challenges that should be addressed in sponsored search. 
The first one is the {\bf keyword bidding problem}: due to the Matthew effect~\cite{merton1968matthew}, the hot ad keywords become too expensive for most of the advertisers, because too many advertisers bid on such keywords. As a result, many advertisers cannot survive in the auction process to get their desired ad impressions. As reported in ~\cite{zhang2014bid}, 55.3\% advertisers have no ad impression, and 92.3\% advertisers have no ad click at all, which is mainly caused by low bid price or improper keywords that they bid. The second issue is {\bf inefficient ad impressions}: a substantial proportion of search queries, which are unpopular yet relevant to many ad keywords, have less competitive ads (46.6\% search queries)
even no ads (41.0\% search queries) on their search result pages as reported in~\cite{zhang2014bid}. 
Because of the two reasons, the expectation of advertisers is not satisfied and the revenue of search engines is also not optimized.

To address these problems, several prior studies have been conducted in keyword generation or suggestion~\cite{fuxman2008using,qiao2015novel,chen2008advertising,abhishek2007keyword,joshi2006keyword}. Most of these studies adopt matching methods based on the word co-occurrence between ad keywords~\cite{chen2008advertising} and queries~\cite{fuxman2008using}. However, these methods tend to suggest popular keywords to advertisers, which will deteriorate the bidding competition. In addition, these approaches cannot suggest novel ad keywords which do not appear in the corpus.

Recently, deep learning technologies have been applied in many natural language tasks, such as machine translation~\cite{sutskever2014sequence}, ad keyword suggestion~\cite{grbovic2015context}, and query rewriting~\cite{he2016learning}. However, it's not trivial to adapt these neural networks to the ad keyword generation task, due to two major challenges. \textbf{First}, the generated ad keywords should be diversified and relevant to the original keywords to cover more user queries, which is not supported by existing neural models applied in keyword and query generation tasks. \textbf{Second}, the generated ad keywords should satisfy many constraints in sponsored search. For instance, to provide relevant yet unexplored ads for users, it is necessary to satisfy the domain constraint which means that the generated keywords should belong to the domain of the source keyword or several appropriate domains. For instance, a keyword in the {\it health care} domain should only match the keywords from the same domain to ensure the ad quality, while a keyword from the {\it information} domain could match those from various domains, such as {\it entertainment} and {\it shopping}, to cover diverse user queries.

In this paper, we investigate to use generative neural networks in the task of ad keyword generation. Given the purchased keyword as input, our generative model can suggest a set of keywords based on the semantics of the input keyword, as shown in Figure \ref{fig:process}. The generated keywords are diverse and even completely novel (the blue keywords) from those in the dataset. This generative approach can address the aforementioned problems in two ways.
{\bf First}, our model is able to generate diverse, novel keywords, instead of merely suggesting the existing popular keywords in the dataset, which can recommend keywords for advertisers to alleviate the keyword bidding problem and retrieve ads by keyword reformulation for sponsored search engines to address the inefficient ad impression issue.
{\bf Second}, to improve the quality of the generated keywords, we incorporate the domain constraint in our model, which is a key factor considered in sponsored search to display ads. Through capturing the domain constraint, our model learns both semantic information and domain-specific information of ad keywords during training, and is consequently able to predict the proper domain category and generate the ad keyword based on the predicted category. 
In addition, our model uses reinforcement learning to strengthen the domain constraint in the generation process, which further improve the domain correlation and the keyword quality.

To summarize, this paper makes the following contributions:
\begin{itemize}

\item This work investigates to use generative neural networks for keyword generation in sponsored search, which addresses the issues of keyword bidding and inefficient ad impressions.

\item We present a novel model that incorporates the domain constraint in ad keyword generation. The model is able to predict a suitable domain category and generate an ad keyword correspondingly. A reinforcement learning algorithm is devised to adaptively utilize domain-specific information in keyword generation, which further improves the domain consistency and the keyword quality.

\item We perform offline and online evaluation with the proposed model, and extensive results demonstrate that our model can generate diverse, novel, relevant, and domain-consistent keywords, and also improves the performance of sponsored search.

\end{itemize}

\section{RELATED WORK}

\subsection{Keyword Generation}
A variety of methods has been proposed for generating and suggesting the keywords for advertisements, as ad keywords play a critical role in sponsored search. \citeauthor{joshi2006keyword} ~\cite{joshi2006keyword} collected text-snippets from search engine given the keyword as input, and constructed them as a graph model to generate relevant keywords based on the similarity score. \citeauthor{abhishek2007keyword}~\cite{abhishek2007keyword} further improved the graph model, which computes the similarity score based on the retrieved documents. \citeauthor{chen2008advertising}~\cite{chen2008advertising} applied concept hierarchy to keyword generation, which suggests new keywords according to the concept information rather than the co-occurrence of the keywords itself. \citeauthor{fuxman2008using}~\cite{fuxman2008using} made use of the query-click graph to compute the keyword similarity for recommendation based on a random walk with absorbing states. \citeauthor{ravi2010automatic}~\cite{ravi2010automatic} introduced a generative approach, a monolingual statistical translation model, to generate bid phrases given the landing page, which performs significantly better than extraction-based methods. Recently due to the advances of deep learning, various neural network models have been applied to ad keyword suggestion. \citeauthor{grbovic2015context}~\cite{grbovic2015context} proposed several neural language models to learn low-dimensional, distributed representations of search queries based on context and content of the ad queries within a search session. \citeauthor{zhai2016deepintent}~\cite{zhai2016deepintent} applied an attention network which is stacked on top of a recurrent neural network (RNN) and learns to assign attention scores to words within a sequence (either a query or an ad).

\subsection{Query Generation}

Another related research topic is query generation, which is widely studied and applied in organic search. It improves user experience by either expanding (reformulating) a user's query to improve retrieval performance, or providing suggestions through guessing the user intention, according to the user's behaviour pattern (query suggestion). Some previous studies adopt query logs to generate queries by handcrafted features such as click-through data~\cite{cui2002probabilistic,zhang2007comparing,leung2008personalized,sloan2015term}, session based co-occurrence~\cite{jones2006generating,he2009web,jiang2014learning} or query similarity~\cite{fonseca2005concept,antonellis2008simrank++,bing2015web}. Recently, artificial neural networks have been applied in query processing. A hierarchical recurrent encoder-decoder model~\cite{sordoni2015hierarchical} is introduced to query suggestion. \citeauthor{he2016learning}~\cite{he2016learning} proposed a learning to rewrite framework consisting of a candidate generating phase and a candidate ranking phase for query rewriting. \citeauthor{song2017translation}~\cite{song2017translation} use an RNN encoder-decoder to translate a natural language query into a keyword query. An attention based hierarchical neural query suggestion model that combines a session-level neural network and a user-level neural network to model the short- and long-term search history of a user is proposed by \citeauthor{chen2018attention}~\cite{chen2018attention}

\subsection{Generative Neural Network}
Recently, generative neural networks have been applied in many natural language tasks, such as machine translation~\cite{sutskever2014sequence}, dialogue generation~\cite{Shang2015Neural}, and query rewriting~\cite{he2016learning}. \citeauthor{sutskever2014sequence}~\cite{sutskever2014sequence} apply an end-to-end approach, a sequence to sequence (Seq2Seq) model, on machine translation tasks. \citeauthor{Shang2015Neural}~\cite{Shang2015Neural} further introduce the Seq2Seq model to dialogue generation tasks with novel attention mechanisms. Although the Seq2Seq model is capable of generating a sequence with a certain meaning, it isn't suitable for diversified sequence generation as argued in~\cite{2016li-SimpleDiverseDecoding}. Therefore, latent variable based models are proposed to address the diversity and uncertainty problem. \citeauthor{serban2017hierarchical}~\cite{serban2017hierarchical} introduce latent variables to a hierarchical encoder-decoder neural network to explicitly model generative processes that possess multiple levels of variability. \citeauthor{zhou2017mojitalk}~\cite{zhou2017mojitalk} propose several conditional variational autoencoders to use emojis to control the emotion of the generated text. \citeauthor{zhao2017learning}~\cite{zhao2017learning} use latent variables to learn a distribution over potential conversational intents based on conditional variational autoencoders, that is able to generate diverse responses using only greedy decoders.

\section{MODEL}

\subsection{Background: Encoder-Attention-Decoder Framework}

We first introduce a general encoder-attention-decoder framework based on sequence-to-sequence (Seq2Seq) learning \cite{sutskever2014sequence}, which is a widely used generative neural network. The encoder and decoder of the Seq2Seq model \cite{sutskever2014sequence} are implemented with GRU \cite{cho2014learning, chung2014empirical}.

The encoder represents an input sequence $X = x_1x_2\cdots x_{n}$ with hidden representations $\bm{H} = \bm{h}_1\bm{h}_2\cdots \bm{h}_n$\footnote{Throughout the paper, a bold character (e.g., $\bm{h}$) denotes the vector representation of a variable (${h}$).}
, which is briefly defined as below:
\begin{equation}
\bm{h}_t = \mathbf{GRU}(\bm{h}_{t-1},\bm{e}(x_t)),
\label{eq:encodersimple}
\end{equation}
where $\bm{e}(x_{t})$ is the embedding of the word $x_t$, and GRU is gated recurrent unit~\cite{cho2014learning}.

The decoder takes as input a context vector $\bm{c}_t$ and the embedding of a previously decoded word $\bm{e}(y_{t-1})$, and updates its state $\bm{s}_t$ using another GRU:
\begin{equation}
\bm{s}_t = \mathbf{GRU}(\bm{s}_{t-1},[\bm{c}_{t-1};\bm{e}(y_{t-1})]) ,
\label{eq:decodersimple}
\end{equation}
where $[\bm{c}_{t-1};\bm{e}(y_{t-1})]$ is the concatenation of the two vectors, serving as input to the GRU network. 
The context vector $\bm{c}_t$ is designed to attend to the key information of the input sequence during decoding, which is a weighted sum of the encoder's hidden states as $\bm{c}_{t-1}=\sum_{k=1}^n \alpha_{k}^{t-1} \bm{h}_k$, and $\alpha_{k}^{t-1}$ measures the relevance between state $\bm{s}_{t-1}$ and hidden state $\bm{h}_k$. Refer to ~\cite{bahdanau2014neural} for more details. 

Once the state vector $\bm{s}_t$ is obtained, the decoder generates a token by sampling from the generation distribution $\bm{o}_t$ computed from the decoder's state $\bm{s}_t$  as follows: 
\begin{eqnarray}
y_t \sim \bm{o}_t & = & P(y_t \mid y_1, y_2,\cdots, y_{t-1}, \bm{c}_t) ,\\
\label{eq:soft}
& = & {\rm softmax}(\mathbf{W_o} \bm{s}_t) .
\end{eqnarray}

\subsection{Task Definition and Overview}

\begin{figure}[!htp]
  \centering
  \includegraphics[width=0.75\linewidth]{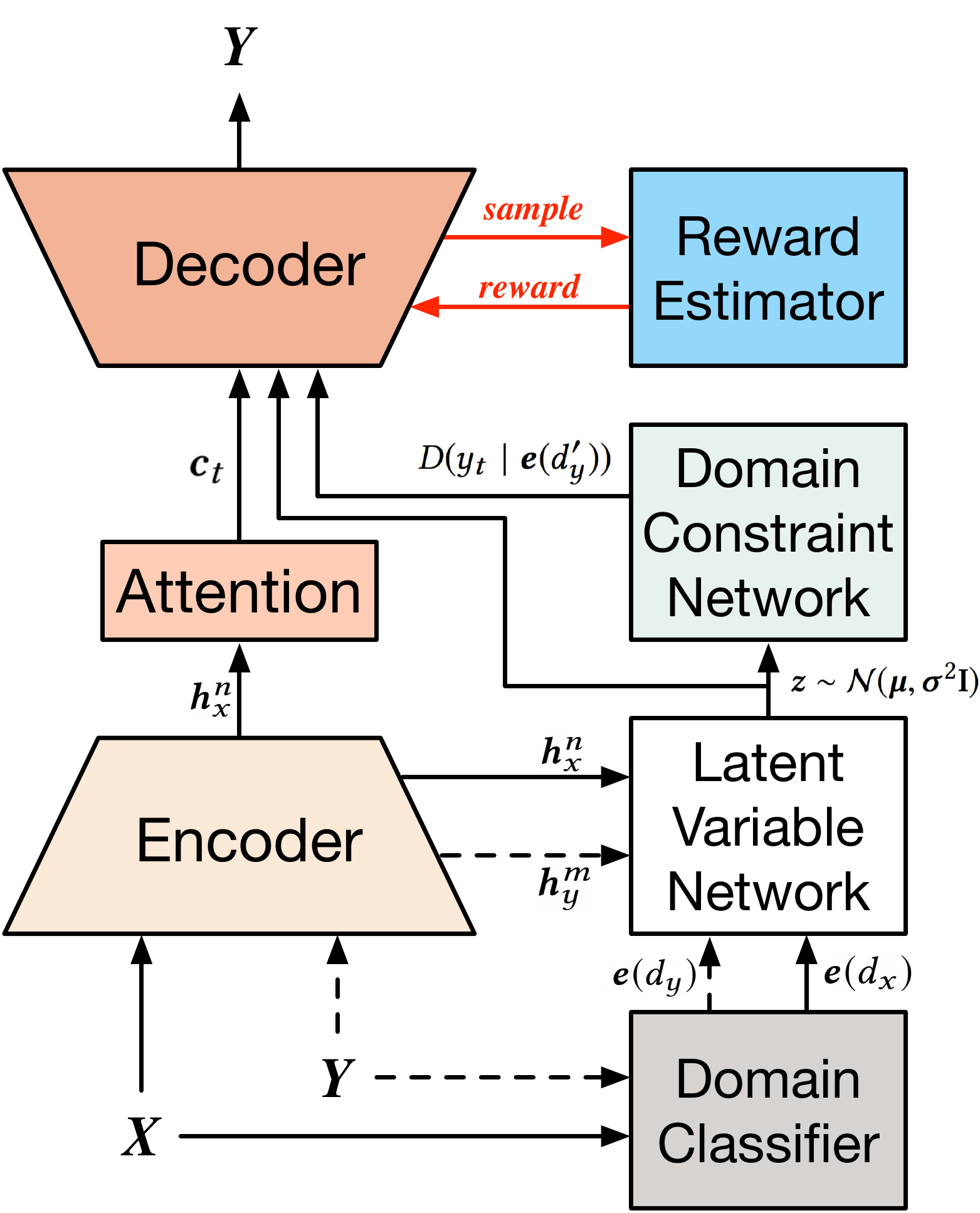}% 1\linewidth
  \caption{Overview of DCKG. The dashed arrow is used only in supervised training. The red arrow is used only in reinforcement learning.
  During reinforcement learning, DCKG infers a set of keyword samples and get their rewards from the reward estimator, which are used to further train our model to improve the quality of generated keywords.}
  \label{fig:overview}
\end{figure}

Our problem is formulated as follows: Given a purchased ad keyword $\bm{X} = (x_1,x_2,\cdots, x_{n})$ (a word sequence)\footnote{Throughout the paper, a keyword refers to a word sequence, but not a single word.} 
and the corresponding domain category $d_x$ obtained from a domain classifier, the goal is to predict a suitable target domain category $d_y$ and generate a target keyword $\bm{Y} = (y_1,y_2,\cdots, y_{m})$ (a word sequence) that is coherent with the domain category $d_y$. Essentially, the model estimates the probability:
$P(\bm{Y}, d_y|\bm{X},d_x)=P(d_y|\bm{X},d_x)\prod_{t=1}^{m}P(y_t|y_{<t},d_y,\bm{X},d_x)$.
The domain categories are adopted from the sponsored search engine, which consists of $k$ domain categories, such as {\it beauty care}, {\it shopping}, and {\it entertainment}.

Building upon the encoder-attention-decoder framework, we propose the Domain-Constrained Keyword Generator (DCKG) to generate diversified keywords with domain constraints using three mechanisms.
{\bf First}, DCKG incorporates a latent variable sampled from a multivariate Gaussian distribution to generate diversified keywords. 
{\bf Second}, a domain constraint network is proposed to facilitate generating domain-consistent keywords, which imposes more probability bias to domain-specific words.
{\bf Third}, DCKG further optimizes the decoder to adjust the word generation distribution with reinforcement learning.

An overview of DCKG is presented in Figure \ref{fig:overview}, which illustrates the dataflow of DCKG in supervised learning, reinforcement learning, and inference processes. In supervised learning, the source keyword $\bm{X} = (x_1,x_2,\cdots, x_{n})$ and the target keyword $\bm{Y} = (y_1,y_2,\cdots, y_{m})$ are fed to the encoder to generate the hidden representations $\bm{h}^n_x$ and $\bm{h}^m_y$, meanwhile, they are fed to the domain classifier to obtain their domain categories $d_x$ and $d_y$ respectively. The domain categories are further converted to the domain embeddings $\bm{e}(d_x)$ and $\bm{e}(d_y)$ to encode the domain-specific information in DCKG. Then, the latent variable $\bm{z}$ is sampled from a multivariate Gaussian distribution, which is determined by a recognition network that takes the hidden representations and the domain embeddings as input. Given the latent variable $\bm{z}$, the hidden representation of source keyword $\bm{h}^n_x$, and the domain embedding $\bm{e}(d_x)$, DCKG predicts the target keyword category $d'_y$ and generate the domain-specific word score $D(y_t \mid \bm{e}(d'_y))$. Finally, the decoder takes as input the context vector $\bm{c}_t$ generated by attention mechanism, the latent variable $\bm{z}$, and the domain-conditioned word score $D(y_t \mid \bm{e}(d'_y))$ to generate the target keyword $\bm{Y}$.

During the inference process, DCKG has the input of only the source keyword $\bm{X} = (x_1,x_2,\cdots, x_{n})$ to generate a target keyword, conditioned on the latent variable sampled from a prior network which is approximated by the recognition network during supervised training. 

During the reinforcement learning process, DCKG first infers a set of keyword samples. Then, these samples are fed to the reward estimator to get their rewards, considering both domain-specific and semantic information. Finally, these rewards are applied to train our model using reinforcement learning to further improve the quality of generated keywords. 

\subsection{Supervised Learning}
\label{model:supervise}

\begin{figure}[!htp]
  \centering
  \includegraphics[width=0.95\linewidth]{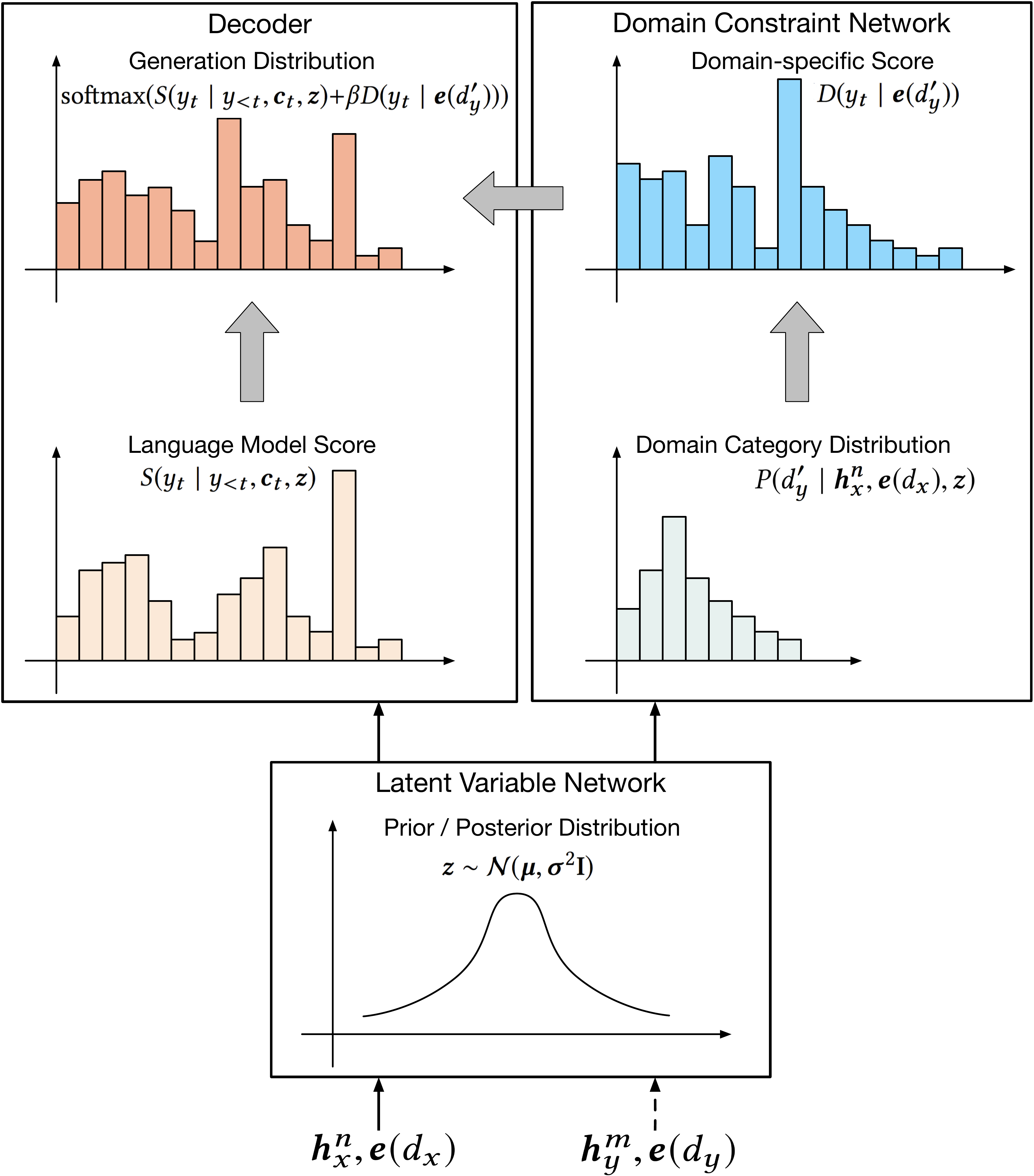}
  \caption{Computation flow of DCKG. The dashed arrow is used only in supervised learning to model the posterior distribution.
  During the inference process, the latent variable $z$ is sampled from the prior distribution, and then fed to the domain constraint network and to the decoder. In the domain constraint network, the latent variable is used to predict the domain category distribution to obtain the target domain category $d'_y$, which is then used to compute the domain-specific score. In the decoder, the latent variable is used to compute the language model score. Finally, the language model score and the domain-specific score are combined to estimate the distribution for word generation.}
  \label{fig:detail}
  \vspace{-4mm}
\end{figure}

We will concentrate on introducing the latent variable network, the domain constraint network, and the decoder network, as shown in Figure \ref{fig:detail}.
The domain classifier is adopted from the sponsored search engine with parameters fixed during training. It generates a one-hot domain category distribution $d_x$ for a keyword. A category is converted to a domain category embedding as follows:
\begin{equation}
\label{eq:domain_embedding}
    \bm{e}(d_x) = \mathbf{V_d}d_x,
\end{equation}
where $\mathbf{V_d}$ is a random initialized domain category embedding matrix which will be learned automatically.

\subsubsection{Latent Variable Network}

Building upon the encoder-attention-decoder framework, our model introduces a latent variable to project ad keywords to a latent space. By this means, the model can generate diversified keywords conditioned on different latent variable sampled from the latent space. Specifically, we adopt the conditional variational autoencoder (CVAE) ~\cite{sohn2015learning} as the latent variable network, which is successfully applied in language generation tasks~\cite{zhou2017mojitalk,zhao2017learning}. 

DCKG assumes the latent variable $\bm{z}$ follows a multivariate Gaussian distribution, $\bm{z} \sim \mathcal{N}(\bm{\mu}, \bm{\sigma}^2\mathbf{I})$. The latent variable network consists of a prior network and a recognition network to model the semantic and domain-specific information of ad keyword in the latent space. 

In the training process, the recognition network takes as input the semantic representations $ \bm{h}_x^n, \bm{h}_y^m $ and the domain embeddings $ \bm{e}(d_x), \bm{e}(d_y) $ of the source keyword $\bm{X}$ and the target keyword $\bm{Y}$, to approximate the true posterior distribution, as $q_\phi(\bm{z} \mid \bm{h}_x^n, \bm{e}(d_x), \bm{h}_y^m, \bm{e}(d_y)) \sim \mathcal{N}(\bm{\mu}, \bm{\sigma}^2\mathbf{I})$. 

In the inference process, the prior network takes only the semantic representation $\bm{h}_x^n$ and the domain embedding $\bm{e}(d_x)$ of the source keyword $\bm{X}$ as input, to sample latent variables from the prior distribution, $p_\theta(\bm{z} \mid \bm{h}_x^n, \bm{e}(d_x)) \sim \mathcal{N}(\bm{\mu}', \bm{\sigma}'^2\mathbf{I})$. The prior/posterior distribution can be parameterized by neural networks such as a multilayer perceptron (MLP) as follows:
\begin{eqnarray}
\left[\,\bm{\mu}\ , \bm{\sigma}^2\ \right] & = & \mathbf{MLP}(\bm{h}_x^n, \bm{e}(d_x), \bm{h}_y^m, \bm{e}(d_y)) , \\
\left[\bm{\mu}', \bm{\sigma}'^2\right] & = & \mathbf{MLP}(\bm{h}_x^n, \bm{e}(d_x)) ,
\end{eqnarray}

To alleviate the inconsistency between the prior distribution and the posterior distribution, we add the KL divergence term to the loss function as follows:
\begin{equation}
\mathcal{L}_1 = KL(q_\phi(\bm{z} \mid \bm{h}_x^n, \bm{e}(d_x), \bm{h}_y^m, \bm{e}(d_y))\,\|\, p_\theta(\bm{z} \mid \bm{h}_x^n, \bm{e}(d_x)))
\end{equation}

\subsubsection{Domain Constraint Network}

The domain constraint network is designed to model the domain-specific information of ad keyword. The output of the network is further incorporated into the process of keyword generation to improve the quality of generated keywords in sponsored search. It plays two roles in our model: first, predicting an appropriate domain category given a latent variable; second, endowing the generated keyword with the target domain features.

Essentially, it predicts a domain category distribution conditioned on the latent variable sampled from a multivariate Gaussian distribution. Once the domain category distribution is determined, we can sample a target domain category by an ${\rm argmax}$ operation. However, ${\rm argmax}$ operation is non-differentiable and the training signal cannot be backpropagated. Inspired by the Gumbel-Max trick~\cite{gumbel1954statistical,maddison2014sampling}, we adopt Gumbel-Softmax~\cite{jang2016categorical} as a differentiable substitute to generate a sample from the domain category distribution, which is defined as follows: 
\begin{eqnarray}
    \bm{\epsilon} & \sim & \mathbf{U}(0, 1),\\
    \bm{g} & = & - {\rm log}(- {\rm log}(\bm{\epsilon})),\\
    \bm{o}_d & = & \mathbf{U_d}\,\mathbf{MLP}(\bm{h}_x^n, \bm{e}(d_x), \bm{z}),\\
    \label{eq:d_y}
    P_{real}(d'_y \mid \bm{h}_x^n, \bm{e}(d_x), \bm{z}) & = & {\rm softmax}(\bm{o}_d),\\
    P_{sample}(d'_y \mid \bm{h}_x^n, \bm{e}(d_x), \bm{z}) & = & {\rm softmax}((\bm{o}_d + \bm{g}) / \tau)
\end{eqnarray}
where $\bm{\epsilon}$ is a sample from the uniform distribution $\mathbf{U}(0, 1)$, $\bm{g}$ is a sample from the Gumbel distribution $\mathbf{Gumbel}(0, 1)$, $\bm{o}_d$ is the logits computed by a MLP and a projection matrix $\mathbf{U_d} \in \mathbb{R}^{k \times n}$; $P_{real}(d'_y \mid \bm{h}_x^n, \bm{e}(d_x), \bm{z})$ is the real distribution of the predicted domain category used in the inference process, $P_{sample}(d'_y \mid \bm{h}_x^n, \bm{e}(d_x), \bm{z})$ is the sample distribution used in the training process, and $\tau$ is a temperature used to adjust the shape of the sample distribution, which is annealed during training.

In supervised learning, we use the ground-truth domain category $d_y$ of a keyword as the supervision signal in the loss function, such that the domain constraint network can predict the target domain category as expected, which is defined as follows:
\begin{equation}
\mathcal{L}_2 = - \mathbf{E}_{q_\phi(\bm{z} \mid \bm{h}_x^n, \bm{e}(d_x), \bm{h}_y^m, \bm{e}(d_y)} [\log P_{real}(d'_y \mid \bm{h}_x^n, \bm{e}(d_x), \bm{z})]
\end{equation}

Another task of the domain constraint network is to compute the domain-specific score of a generated word from the domain category distribution. The domain-specific score is added to the word generation distribution in the decoder to endow a generated keyword with desirable domain-specific features.
When the target domain category distribution is obtained, the target domain embedding can be computed as follows:
\begin{equation}
    \bm{e}(d'_y) = \mathbf{V_d}P(d'_y \mid \bm{h}_x^n, \bm{e}(d_x), \bm{z}),
\end{equation}
where $\mathbf{V_d}$ is the domain category embedding matrix as introduced in Eq. \ref{eq:domain_embedding}. Subsequently, taking the target domain embedding as input, the domain word score is generated as follows:
\begin{equation}
    D(y_t \mid \bm{e}(d'_y)) = \mathbf{W_d}\,\mathbf{MLP}(\bm{e}(d'_y)),
\end{equation}
where $D(y_t \mid \bm{e}(d'_y))$ is the domain-specific score of a generated word, which models the domain-specific features of a target keyword, $\mathbf{W_d}$ is the domain word embedding matrix.

\subsubsection{Decoder Network}
The decoder of DCKG incorporates the latent variable and the domain word score to generate an ad keyword. Taking as input the latent variable $\bm{z}$ and the context vector $\bm{c}_t$, the language model score of a generated word, which captures the semantic information in a keyword, is generated as follows: 
\begin{eqnarray}
    \bm{s}_t & = & \mathbf{GRU}(\bm{s}_{t-1},[\bm{c}_{t-1};\bm{z};\bm{e}(y_{t-1})]),\\
    S(y_t \mid y_{<t}, \bm{c}_t, \bm{z}) & = & \mathbf{W_s} \bm{s}_t,
\end{eqnarray}
where $S(y_t \mid y_{<t}, \bm{c}_t, \bm{z})$ is the language model score and $\mathbf{W_s}$ is the semantic word embedding matrix.

Finally, the decoder combines the language model score and the domain-specific score with a factor $\beta$ and then normalizes the result to the word generation distribution, which is defined as follows:
\begin{equation}
    P(y_t \mid y_{<t}, \bm{c}_t, \bm{z}, \bm{e}(d'_y)) = {\rm softmax}(S(y_t | y_{<t}, \bm{c}_t, \bm{z}) + \beta D(y_t | \bm{e}(d'_y))),
\end{equation}
where $\beta$ is a domain constraint factor that controls the influence of the domain-specific information in the final word generation distribution, and is fixed to 1.0 during supervised training.
The generation loss of the decoder is given as below:
\begin{equation}
\mathcal{L}_3 = - \mathbf{E}_{q_\phi(\bm{z} \mid \bm{h}_x^n, \bm{e}(d_x), \bm{h}_y^m, \bm{e}(d_y)} [\sum_{t=1}^m\log P(y_t \mid y_{<t}, \bm{c}_t, \bm{z}, \bm{e}(d'_y))].
\end{equation}

\subsubsection{Loss Function}
The final loss to be minimized in supervised learning is the combination of the KL divergence term $\mathcal{L}_1$, the domain prediction loss $\mathcal{L}_2$, and the generation loss $\mathcal{L}_3$:
\begin{equation}
\mathcal{L} = {\rm max}(\delta, \mathcal{L}_1) + \mathcal{L}_2 + \mathcal{L}_3,
\end{equation}
where the max operation and the factor $\delta$ are used to balance the KL divergence term and other loss function for better optimization, which is known as the free bits method in~\cite{kingma2016improved}.

\subsection{Reinforcement Learning}
\label{sec:reinforce}

One major disadvantage of DCKG described above is the domain constraint factor $\beta$ is fixed for all the keywords in any domain. However, the optimal factor should be determined by the semantic and domain-specific information of a keyword dynamically. A lower $\beta$ value leads to keywords containing less domain-specific features, while a higher value results in keywords that are less fluent or relevant but contain more domain-specific features, as shown in Section \ref{sec:auto}. Therefore, we propose a reinforcement learning algorithm that is able to learn different $\beta$ values for different keywords.

\subsubsection{Policy Network}
To explore suitable $\beta$ values for different keywords, we first define a value space $\mathcal{B}$ which contains feasible $\beta$ values. The main idea is to choose the best $\beta$ value by a policy network to achieve the maximum reward with respect to the evaluation metrics, which can be implemented in three steps:
first, generate a set of keywords with different $\beta$ values sampled from $\mathcal{B}$, given the same source keyword and latent variable; second, obtain the reward of each keyword using the reward estimator; third, update the policy network to choose the $\beta$ value that leads to a maximum reward.
The policy network $\pi_\psi(\beta \mid \bm{X}, \bm{z})$, parameterized by $\psi$, is formally given below:
\begin{eqnarray}
    \bm{o}_b & = & [\bm{h}_x^n; \bm{e}(d_x); \bm{z}; \bm{e}(d'_y)], \\
    \pi_\psi(\beta \mid \bm{X}, \bm{z}) & = & {\rm softmax}(\mathbf{W_b}\, \mathbf{MLP}(\bm{o}_b)),
\end{eqnarray}
where $\mathbf{W_b} \in \mathbb{R}^{b \times n}$ is a matrix projecting the input vector $\bm{o}_b$ to the action space $\mathcal{B}$. 

We use the REINFORCE algorithm~\cite{williams1992simple}, a policy gradient method, to optimize the parameters by maximizing the expected reward of a generated keyword as follows:
\begin{equation}
    \mathcal{J}(\psi) = \mathbb{E}_{\beta \sim \pi_\psi(\beta \mid \bm{X}, \bm{z})}[R(\bm{X}, \bm{z}, \beta)],
\end{equation}
where $R(\bm{X}, \bm{z}, \beta)$ is the normalized reward of a generated keyword. 
It is noteworthy that the policy network cannot be optimized through supervised learning, as the ground-truth $\beta$ value is not observable.

\subsubsection{Reward Estimator}
The reward estimator is designed to provide a reward balancing the domain-specific information and the semantic information for a generated keyword. To estimate the reward, given the source keyword $\bm{X}$, the latent variable $\bm{z}$, and the target domain category $d'_{y}$ as the same input, we first sample a set of $\beta$ values $\{\beta_{1},\beta_{2},\cdots,\beta_{k}\}$, and infer a set of keyword samples $\{\bm{Y}_1,\bm{Y}_2,\cdots,\bm{Y}_k\}$ based on different $\beta$ samples using the proposed model. Then, we use the domain classifier to predict the domain category $d_{y_i}$ of each keyword $\bm{Y}_i$. The agreement $\gamma$ between $d'_{y}$ and $d_{y_i}$ is treated as an evaluation metric of the domain-specific information, which is defined as follows:
\begin{equation}
    \label{eq:gamma}
    \gamma_i = 
    \begin{cases} 
    1,  & \mbox{if }d'_{y} = d_{y_i} \\
    0, & \mbox{if }d'_{y} \neq d_{y_i} 
    \end{cases}
\end{equation}
Subsequently, we use the generation probabilities computed by a language model and DCKG as an evaluation metric of the semantic information. Finally, the min-max normalization is applied to re-scale the rewards $\{{r_1},{r_2},\cdots,{r_k}\}$ for each $\beta_i$ and $\bm{Y}_i$. This process is defined as follows:
\begin{eqnarray}
    \label{eq:reward}
    r_i & = & \gamma_i (\lambda P_{LM}(\bm{Y}_i) + (1-\lambda) P_{DCKG}(\bm{Y}_i)),\\
    R_i(\bm{X}, \bm{z}, \beta_i) & = & \frac{r_i - {\rm min}(\{r_i\})}{{\rm max}(\{r_i\}) - {\rm min}(\{r_i\})},
\end{eqnarray} 
where $r_i$ is the reward for each $\beta_i$ and $\bm{Y}_i$, $\gamma_i$ is the domain category agreement, $P_{LM}$ and $P_{DCKG}$ are the generation probabilities modeled by a language model and our DCKG respectively, $\lambda \in (0, 1)$ is a weight to balance these two generation probabilities. 

And the gradient is approximated using the likelihood ratio trick~\cite{glynn1990likelihood} as:
\begin{equation}
    \nabla \mathcal{J}(\psi) = R(\bm{X}, \bm{z}, \beta) \nabla {\rm log} \pi_\psi(\beta \mid \bm{X}, \bm{z}),
\end{equation}

This RL component encourages the model to choose for each keyword an optimal factor $\beta$ that will lead to both higher domain category agreement and generation quality. Through reinforcement learning, we can further improve the quality of keywords generated by DCKG, resulting in more relevant, fluent, and domain-specific keywords.

\section{EXPERIMENTS}

\subsection{Dataset}

\begin{table} [!htp]
\centering
\begin{tabular}{|c|c|c|c|c|}
\hline
\multicolumn{2}{|c|}{Pairs} & Details & Keywords & Queries\\
\hline
\multirow{2}{*}{Training} & \multirow{2}{*}{43,756,585} & Number & 4,419,555 & 19,203,972\\
\cline{3-5}
&& Length & 3.96 & 5.40\\
\hline
\multirow{2}{*}{Validation} & \multirow{2}{*}{10,000} & Number & 9,504 & 9,997\\
\cline{3-5}
&& Length & 3.73 & 5.21\\
\hline
\multirow{2}{*}{Test} & \multirow{2}{*}{10,000} & Number & 9,474 & 9,996\\
\cline{3-5}
&& Length & 3.74 & 5.21\\
\hline
\end{tabular}\caption{Statistics of the dataset.}
\label{tab:dataset}
\vspace{-6mm}
\end{table}

We sampled about 40 million query logs from Sogou.com, and each sample consists of a <{\it ad keyword, user query}> pair. When the search engine retrieves ads based on their owners' purchased keywords, the ad keyword must be relevant to the user query.
Therefore, we treat user queries as target keywords to train our model. In other words, the input to our model is a purchased ad keyword ($\bm{X}$), and the expected output is a user query ($\bm{Y}$). 
We randomly sampled 10,000 pairs for validation and test. The statistics are presented in Table \ref{tab:dataset}.

\subsection{Implementation Details}

Our model was implemented with Tensorflow~\cite{abadi2016tensorflow}. The encoder and decoder do not share parameters and each has 4-layer GRU networks with 1024 hidden neurons in each layer. The word embedding size is set to 1,024. The vocabulary size is limited to 40,000. The MLP is implemented as a one-layer linear transformation with a ${\rm tanh}$ activation function. The parameters of the MLP and the embedding matrix share the same hidden size $n = 1,024$. All the parameters and embeddings are randomly initialized and tuned during end-to-end training. 

We adopted top $k = 25$ frequent domain categories produced by the domain classifier, which is a key component used in the sponsored search engine. The domain classifier is a support vector machine~\cite{joachims1998making} trained on a large human-labeled corpus.
The accuracy of the $25$-class classification is 92.65\%. 
The temperature $\tau$ is set to  3.0 at the beginning, and is annealed to 0.1 during supervised learning. The factor $\delta$ is set to 5.0. The value space $\mathcal{B}$ consists of $b = 21$ values in the range $[0, 5]$ with an interval of 0.25. The generation probability factor $\lambda$ is set to 0.9.

Our model is trained in two stages: supervised learning is first used to model the semantic information and the domain-specific information of keywords; reinforcement learning is then used to adaptively utilize domain-specific information in the generation process. 
We used the Adam optimizer with a mini-batch size of 100. The learning rate is 0.0001 for supervised learning and 0.00001 for reinforcement learning. The models were run at most 10 epochs for supervised learning and 2 epochs for reinforcement learning. The training process of each model took about a week on a Tesla V100 GPU machine.

\begin{table*} [!htp]
\centering
\begin{tabular}{l|l|l|l|l|l|l|l|l|l|l|l|l}
\hline
\multirow{2}{*}{Model} & \multicolumn{2}{c|}{Perplexity} & \multicolumn{2}{c|}{Perplexity$_{LM}$} & \multicolumn{2}{c|}{Accuracy} & \multicolumn{2}{c|}{Distinct-2} &
\multicolumn{2}{c|}{Distinct-3} &
\multicolumn{2}{c}{Distinct-4} \\
\cline{2-13}
& avg. & std. & avg. & std. & avg.(\%) & std.(\%) & avg.(\%) & std.(\%) & avg.(\%) & std.(\%) & avg.(\%) & std.(\%) \\
\hline
Seq2Seq & 18.82 & 0.033 & 12.26 & 0.022 & 76.84 & 0.054 & 39.09 & 0.028 & 46.78 & 0.033 & 53.02 & 0.041\\
CVAE & {\bf 10.76} & 0.021 & {\bf 8.24} & 0.023 & 77.40 & 0.057 & 69.58 & 0.051 & 83.04 & 0.060 & 90.32 & 0.065\\
\hline
DCKG & 11.21 & 0.025 & 8.94 & 0.034 & {\bf 84.61} & 0.080 & {\bf 71.07} & 0.077 & {\bf 84.26} & 0.065 & {\bf 91.26} & 0.051\\
\hline
\end{tabular}
\caption{Automatic evaluation.}
\label{tab:auto_res}
\vspace{-6mm}
\end{table*}

\begin{table*} [!htp]
\centering
\begin{tabular}{l|l|l|l|l|l|l|l|l|l|l|l|l}
\hline
\multirow{2}{*}{Model} & \multicolumn{2}{c|}{Novelty$_{ALL}$-2} & \multicolumn{2}{c|}{Novelty$_{ALL}$-3} & \multicolumn{2}{c|}{Novelty$_{ALL}$-4} & \multicolumn{2}{c|}{Novelty$_{AD}$-2} &
\multicolumn{2}{c|}{Novelty$_{AD}$-3} &
\multicolumn{2}{c}{Novelty$_{AD}$-4} \\
\cline{2-13}
& avg. & std. & avg. & std. & avg.(\%) & std.(\%) & avg.(\%) & std.(\%) & avg.(\%) & std.(\%) & avg.(\%) & std.(\%) \\
\hline
Seq2Seq & 0.54 & 0.005 & 5.33 & 0.015 & 15.53 & 0.020 & 2.65 & 0.010 & 12.40 & 0.026 & 24.73 & 0.019 \\
CVAE & 2.01 & 0.016 & 16.03 & 0.058 & 38.38 & 0.119 & 8.18 & 0.024 & 33.49 & 0.085 & 57.64 & 0.099 \\
\hline
DCKG & {\bf 2.60} & 0.021 & {\bf 18.23} & 0.056 & {\bf 41.48} & 0.114 & {\bf 9.34} & 0.035 & {\bf 36.01} & 0.095 & {\bf 60.40} & 0.106 \\
\hline
\end{tabular}
\caption{Novelty evaluation.}
\label{tab:novelty_res}
\vspace{-6mm}
\end{table*}

\subsection{Baselines}

As aforementioned, this paper is the first work to address the domain constraint factor in ad keyword generation. We did not find closely-related baselines in the literature. 

Although retrieval-based approaches proposed in previous studies are able to recommend relevant keywords from the dataset, these methods cannot generate novel keywords and thus deteriorate the bidding competition problem. Furthermore, as the dataset consists of <{\it ad keyword, user query}> pairs, retrieval-based methods can only suggest keywords which are already covered by the search engine. Consequently, the coverage and revenue of the search engine cannot be improved more than those methods which can generate novel keywords. Due to these reasons, we didn't consider retrieval-based methods as our baselines.

Nevertheless, we chose two suitable generative models as our baselines: 
\begin{itemize}
\item A modified Seq2Seq model~\cite{sutskever2014sequence}. First, the target domain category is predicted by the domain constraint network and is embedded into a category vector. Then, the vector serves as input to each decoding position.

\item A modified CVAE model~\cite{sohn2015learning} considering the domain constraint information, which is implemented similarly to DCKG but without reinforcement learning.
\end{itemize}

\subsection{Automatic Evaluation}
\label{sec:auto}

We evaluated the quality of keywords generated by different models using several automatic metrics. 
To evaluate the performance in metrics except for perplexity\footnote{We generated the ground-truth target keyword using greedy search for calculating the perplexity.}, we generated 10 keywords for each purchased ad keyword using different models (for DCKG and CVAE, keywords were generated conditioned on 10 sampled latent variables using greedy search; for Seq2Seq, keywords were inferred using beam search with $beam size = 10$).
Due to the uncertainty caused by the latent variable in our model and the CVAE baseline, we repeatedly tested all the models for 10 times. The averaged performance, the standard deviation, and the significance test results were reported for these models \footnote{For the deterministic Seq2Seq model, we tested 10 models which achieve the top 10 performance in the validation set. These models differ in different training steps.}. 
The bold number indicates the best performing model, which outperforms all other models significantly (2-tailed t-test, $p-value < 0.01$).

\subsubsection{Metrics}
\begin{itemize}
\item Perplexity: We adopted perplexity to evaluate the generation quality with respect to grammar and fluency. The perplexity here is computed by the generation distribution in the models themselves. 
\item Perplexity$_{LM}$: We trained a language model on our dataset, and used this language model to calculate the perplexity of keywords generated by different models.
\item Accuracy: To evaluate whether the model can generate domain-constrained keywords, we adopted the domain accuracy as the agreement between the expected domain category (as predicted by the model, see Eq. \ref{eq:d_y}) and the domain category of a generated keyword predicted by the domain classifier. 
\item Distinct-$n$: We calculated the proportion of distinct $n$-grams to all the $n$-grams in generated keywords to evaluate the diversity.
\item Novelty$_{ALL/AD}$-n: To evaluate the novelty of the generated keywords, we counted the distinct $n$-grams that don't appear in all the corpus (ALL) or not in the purchased ad keyword set (AD). Novelty$_{ALL}$-n means the percentage of totally novel $n$-grams that have not been observed in the corpus. 
\end{itemize}

\subsubsection{Results}

The results are shown in Table \ref{tab:auto_res} and Table \ref{tab:novelty_res}, from which we made the following observations:

{\bf First}, DCKG obtains the best performance in domain accuracy, distinct, and novelty. The improvement in domain accuracy shows that our model can enhance the domain consistency of the generated keywords, thanks to the reward we applied in reinforcement learning.
DCKG can generate more diverse and novel keywords than baselines, which can be attributed to the latent variable network we adopted and the domain constraint we considered. Our model has better perplexity and perplexity$_{LM}$ than Seq2Seq but slightly worse than CVAE, because DCKG is tuned to satisfy the domain constraint more by reinforcement learning than fit to the language generation probability. Nevertheless, the disadvantage in perplexity and perplexity$_{LM}$ harms the grammaticality of the generated keywords but less influences the overall performance, as shown in the following experiments. 

{\bf Second}, the novelty scores indicate that our model can generate more novel $n$-grams than the baselines. The generative models can generate novel keywords which have not been purchased yet so that the bidding competition problem can be alleviated as aforementioned. For retrieval-based methods, the Novelty$_{ALL}$ scores, which count $n$-grams that are never observed in the corpus, will be zero since they can only suggest existing keywords from the corpus.

\begin{figure}[!htp]
  \centering
  \includegraphics[width=1.0\linewidth]{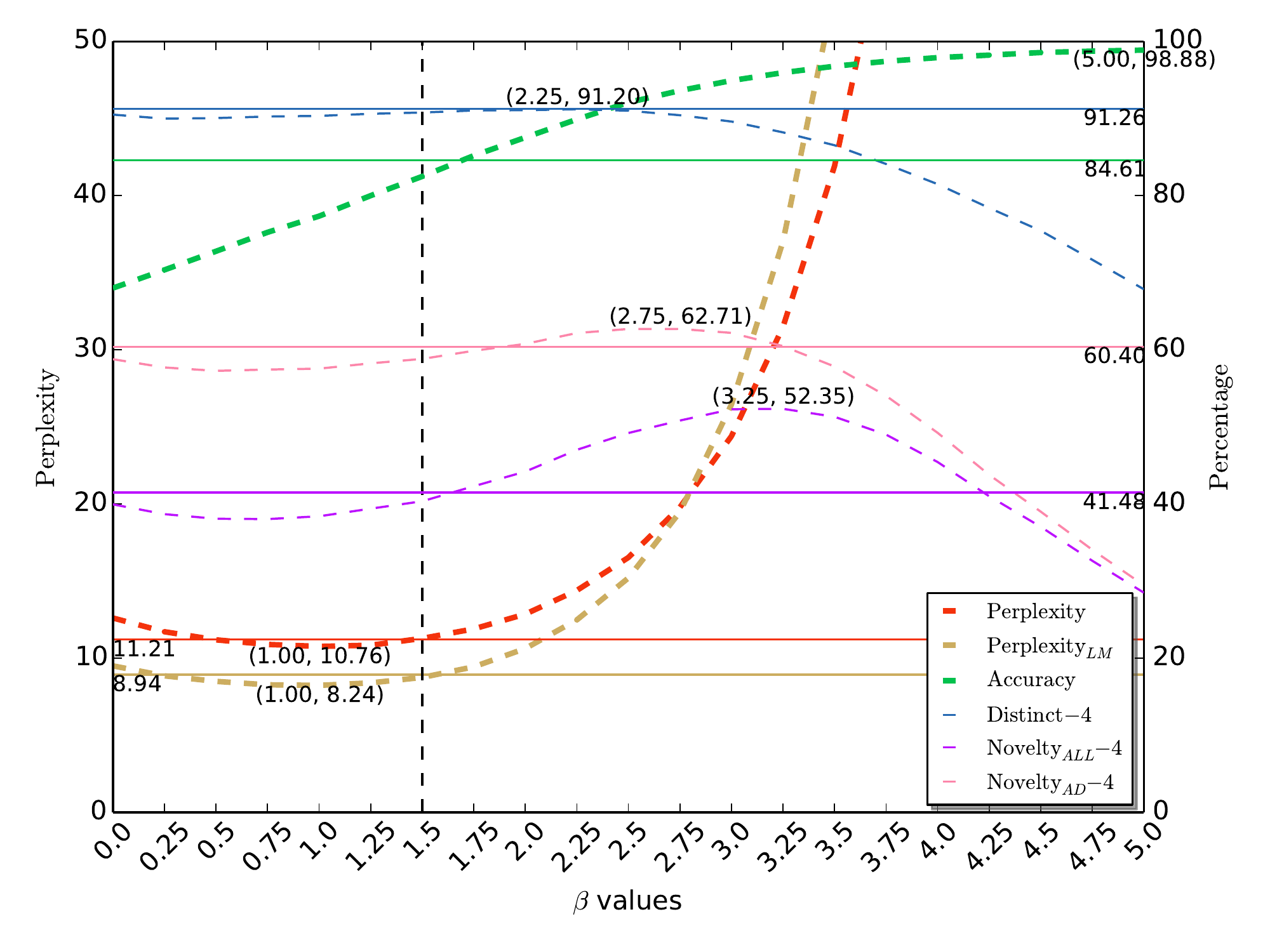}% 1\linewidth
  \caption{Performance curve with varying $\bm{\beta}$. The dashed curve is the performance of DCKG with manually fixed $\bm{\beta}$ in different metrics. 
  The bold curves (perplexity, perplexity$_{\bm{LM}}$, accuracy) show the change of rewards which are used reinforcement learning (see Eq. \ref{eq:reward}). 
  The horizontal line is the performance of DCKG with dynamic $\bm{\beta}$ tuned by RL. 
  The dashed vertical line is an auxiliary line to compare DCKG with fixed $\bm{\beta = 1.5}$ to DCKG with dynamic $\bm{\beta}$ optimized by RL. Left axis: Perplexity and Perplexity$_{\bm{LM}}$. Right axis: Accuracy, Distinct-4, Novelty$_{\bm{ALL}}$-4, and Novelty$_{\bm{AD}}$-4.}
  \label{fig:beta_res}
  \vspace{-4mm}
\end{figure}

\subsubsection{Effect of $\beta$}

In order to investigate the influence of the domain constraint factor $\beta$ in DCKG, we tested the performance of DCKG with different fixed $\beta \in \mathcal{B}$. The results are shown in Figure \ref{fig:beta_res}. As we can see, the domain accuracy of DCKG improves continuously with the increase of $\beta$ (the green curve), indicating that $\beta$ is critical to satisfying the domain constraint in keyword generation because larger $\beta$ applies more domain-specific information in keyword generation. 
DCKG achieves the best performance in perplexity and perplexity$_{LM}$ with $\beta = 1$ since this value is the default $\beta$ value during supervised learning. However, when $\beta$ becomes larger, DCKG performs worse in the perplexity metrics.
It is thus important to balance the domain constraint and the language generation quality. Therefore, we apply reinforcement learning to find an optimal $\beta$ for different instances dynamically. 

It is noteworthy that DCKG with the dynamically determined $\beta$ outperforms DCKG with the manually adjusted $\beta$. For example, although the performance of DCKG in perplexity and perplexity$_{LM}$ is approximately equal to DCKG with the fixed $\beta = 1.5$, the domain accuracy of DCKG outperforms DCKG with the fixed $\beta = 1.5$ ({\bf 84.61} vs. 82.49). Moreover, the novelty score and the diversity score of DCKG are also better than DCKG with the fixed $\beta = 1.5$, for instance, the Distinct-4 of DCKG is better than DCKG with any fixed $\beta \in \mathcal{B}$ ({\bf 91.26} vs. 91.20), indicating that the domain constraint can further improve the quality of generated keywords.

\subsection{Manual Evaluation}
\label{sec:manual}

In manual evaluation, we tested whether the generated keywords are relevant to the source keyword and consistent with the expected domains, such that they are suitable to be suggested to advertisers. 
We manually evaluated the relevance and the grammaticality of keywords generated by different models. We randomly sampled 100 purchased ad keywords from the test set, and generated 10 keywords for each purchased ad keyword using different models as in automatic evaluation.
Five experienced annotators were recruited to judge the relevance ({\it Relevant/Irrelevant}) of 3000 pairs of generated and purchased keywords. The grammaticality of a generated keyword is also considered during annotation, as bad grammaticality leads to the ambiguity in the meaning and thus results in irrelevance rating. We used majority voting to obtain the final label of annotation. 
Moreover, as domain accuracy is critical to measuring how well the models satisfy domain constraint, we also reported the automatic measure in these tables. 
The bold number indicates the best performing model, which outperforms all other models significantly (2-tailed sign test, $p-value < 0.05$).

\subsubsection{Metrics}
\begin{itemize}
\item Precision (P): Precision is defined as the ratio of the number of relevant
keywords generated to the number of total keywords generated. 

\item Recall (R): Recall is the proportion of relevant
keywords generated by a model to all the relevant keywords pooled from all models' results. Similar to \cite{joshi2006keyword}, the relevant keywords from all the models are pooled together and are treated as the entire relevant set. This metric is useful to compare the ability to cover the potential relevant keywords.

\item Accuracy (A): Similar to the preceding section, this metric is the agreement between the expected domain category, predicted by the model, and the domain category of a generated keyword, predicted by the domain classifier (SVM). 

\item F-measures: We adopted four F-measures to quantify the overall performance including F(PR), F(PA), F(RA), and F(PRA). Each F-measure is defined as the harmonic mean of all factors, where P/R/A indicate precision/recall/accuracy respectively.
\end{itemize}

\begin{table} [!htp]
\centering
\begin{tabular}{l|l|l|l}
\hline
Model & Seq2Seq & CVAE$\,\,\,\,$ & DCKG$\,\,\,\,$ \\
\hline
Precision & 0.948 &  0.954 & {\bf 0.970} \\
Recall & 0.325 & 0.332 & {\bf 0.343} \\
Accuracy & 0.761 & 0.774 & {\bf 0.847} \\
\hline
F(PR) & 0.483 & 0.491 & {\bf 0.503} \\
F(PA) & 0.808 & 0.832 & {\bf 0.892} \\
F(RA) & 0.440 & 0.454 & {\bf 0.480} \\
F(PRA) & 0.534 & 0.548 & {\bf 0.574} \\
\hline
\end{tabular}
\caption{Manual evaluation.}
\label{tab:manual_result}
\vspace{-6mm}
\end{table}

\begin{table*} [!htp]
\centering
\begin{tabular}{l|l|l|l|l|l|l|l|l|l}
\hline
\multirow{2}{*}{Day} & \multicolumn{3}{c|}{Seq2Seq} & \multicolumn{3}{c|}{CVAE} & \multicolumn{3}{c}{DCKG}\\
\cline{2-10}
& COV(\%) & CTR(\%) & RPM(\%) & COV(\%) & CTR(\%) & RPM(\%) & COV(\%) & CTR(\%) & RPM(\%) \\
\hline
1 & 39.51 & 44.92 &	21.96 &	53.64 &	59.14 &	24.86 &	58.52 &	63.26 &	30.52 \\
2 & 44.30 &	49.99 &	24.99 &	58.22 &	60.38 &	28.49 &	55.16 &	60.88 &	33.49 \\
3 & 46.38 &	50.02 &	22.52 &	51.53 &	54.04 &	30.30 &	58.07 &	58.78 &	28.24 \\
4 & 47.32 &	46.22 &	16.78 &	51.12 &	57.54 &	34.64 &	58.24 &	61.68 &	28.23 \\
5 & 46.41 &	50.25 &	20.56 &	49.82 &	52.58 &	21.72 &	59.40 &	60.85 &	33.78 \\
6 & 45.27 &	47.06 &	17.36 &	51.61 &	51.53 &	24.95 &	55.20 &	58.17 &	21.68 \\
7 & 41.09 &	45.37 &	26.00 &	51.54 &	55.39 &	24.12 &	54.36 &	59.58 &	24.81 \\
8 & 36.90 &	49.98 &	17.42 &	51.53 &	55.96 &	23.15 &	53.38 &	58.29 &	27.41 \\
9 & 35.89 &	42.85 &	19.25 &	51.92 &	61.53 &	27.00 &	50.60 &	63.24 &	37.75 \\
10 & 39.46 &	49.36 &	25.07 &	56.17 &	60.54 &	28.06 &	53.86 &	57.61 &	22.06 \\
\hline
avg. & 42.25 &	47.60 &	21.19 &	52.71 &	56.86 &	26.73 &	{\bf 55.68} &	{\bf 60.23} &	{\bf 28.80} \\
\hline
\end{tabular}
\caption{Online evaluation. The number reported in each metric is the increased proportion compared to the original keyword reformulation method of the sponsored search engine, which retrieves about 2\% unique ads.}
\label{tab:online_result}
\vspace{-6mm}
\end{table*}

\subsubsection{Annotation Statistics}
We calculated the statistics to measure inter-rater agreement.  
For relevance rating, the percentage of the pairs that all 5 judges gave the same label is 85.1\%, the percentage of the pairs that 4 judges gave the same label (4/5 agreement) amounts to 8.8\%, and the percentage for 3/5 agreement is 6.1\%.
We further computed the free-marginal kappa \cite{randolph2005free} to measure inter-rater consistency. The free-marginal kappa is 0.856, indicating adequate inter-rater agreement.

\subsubsection{Results}
The results are shown in Table \ref{tab:manual_result}. As it can be seen, DCKG obtains the best performance in all the metrics, indicating that our model can generate relevant and grammatical keywords. The best precision shows that more relevant keywords are generated by DCKG. Our model also achieves the best recall, indicating that DCKG is able to cover more potential user queries than other models.
In addition, DCKG outperforms other models in domain accuracy, indicating that incorporating the domain constraint appropriately can enhance the quality of generated keywords.
Moreover, DCKG ranks highest for all the four F-measures, showing that our model is superior considering all these factors.

\subsection{Online Evaluation}
\label{sec:online}
In online evaluation, we examined whether the sponsored search engine can retrieve and present more relevant ads using the keywords generated by our model. We applied the generative models in the keyword reformulation method to facilitate the ad retrieval process of sponsored search (see Figure \ref{fig:process}). Specifically, we collected 5 million purchased ad keywords, and then used each model to generate 10 keywords for each purchased keyword. We used these generated keywords as the reformulated keywords to retrieve original ads. 
The original keyword reformulation method, based on handcrafted rules and templates, is one of the retrieval methods in the ad retrieval process of the sponsored search engine, which retrieves about 2\% unique ads. We added the keywords generated by our models to the keyword reformulation method of the sponsored search engine, which is called the enhanced keyword reformulation method.
To make a fair comparison, the enhanced keyword reformulation methods of all models run ten days in our A/B online test system on Sogou.com, where 10\% user queries are selected into the test system. We compared the performance of the enhanced keyword reformulation method with the original keyword reformulation method, and reported the relative gain in percentage for all the metrics as the final result. The bold number indicates the best performing model, which outperforms all other models significantly (2-tailed t-test, $p-value < 0.05$). 

\subsubsection{Metrics}
\begin{itemize}
\item Coverage (COV): Coverage is defined as the percentage of web pages that contain at least one ad.
\item Click-through rate (CTR): Click-through rate is the ratio of page views that lead to a click to the total number of page views.
\item Revenue per mille (RPM): Revenue per mille is the revenue of the search engine per one thousand page views. %%%
\end{itemize}

\subsubsection{Results}
The results are shown in Table \ref{tab:online_result}. As it can be seen, DCKG obtains the best performance in all the metrics. Results show that keywords generated by DCKG can cover more user queries than those by other models. The click-through rate scores indicate that the new ads which are matched with ad keywords from DCKG lead to more user clicks, because the generated keywords (user queries) are of higher quality and more relevant. The revenue is improved because the click-through rate increases, which verifies that our model can contribute more ad impressions in sponsored search.

Although the Seq2Seq model can also retrieve more ads, the improvements in all metrics are the lowest in the three models, as the diversity and the quality of the generated keywords are lower than ours. CVAE outperforms Seq2Seq because CVAE is able to model diversity with latent variables. However, as CVAE incorporate the domain constraint in a fixed way, it cannot balance the semantic information and the domain-specific information in the generation process. DCKG further improves the diversity and the quality of generated keywords by tuning the $\beta$ factor dynamically with reinforcement learning. The results of online evaluation agree with those of automatic evaluation, indicating that DCKG can generate the most diverse, novel and domain consistent keywords among the three models.

\subsection{Case Study}
\label{sec:case}
\begin{CJK}{UTF8}{gbsn}

\begin{table*} [!htp]
\centering
\small
\begin{tabular}{|c|c|c|l|l|c|c|}
\hline
\multirow{2}{*}{Model} & \multicolumn{2}{c|}{Domain} & \multicolumn{2}{c|}{Keyword} & \multirow{2}{*}{Acc.} & \multirow{2}{*}{Rel.} \\
\cline{2-5}
& \multicolumn{1}{c|}{Original} & \multicolumn{1}{c|}{Translated} & \multicolumn{1}{c|}{Original} & \multicolumn{1}{c|}{Translated} & & \\
\hline
Source &  美容保健 & beauty care & 减肥减肥 & weight loss weight loss & - & - \\
\hline
\multirow{6}{*}{Seq2Seq} 
& 美容保健 & beauty care & 抖音上的减肥的舞蹈视频 & dance video on weight loss & 0 & 1 \\
\cline{2-7}
& 美容保健 & beauty care & 抖音上的减肥的舞蹈视频教程 & dance video tutorial on weight loss & 1 & 1 \\
\cline{2-7}
& \multirow{2}{*}{美容保健} & \multirow{2}{*}{beauty care} & 减肥的人可以吃什么水果 & what fruit is not easy for people & \multirow{2}{*}{1} & \multirow{2}{*}{0} \\
& & & 不容易 &  who want to lose weight & & \\
\cline{2-7}
& \multirow{2}{*}{美容保健} & \multirow{2}{*}{beauty care} & 减肥的人可以吃什么水果 & what fruit is not easy to fat for & \multirow{2}{*}{1} & \multirow{2}{*}{1} \\
& & & 不容易胖 &  people who want to lose weight & & \\
% \cline{2-7}
% & \multirow{2}{*}{美容保健} & \multirow{2}{*}{beauty care} & 减肥的人可以吃什么水果 & what fruit is not easy to be fat for& \multirow{2}{*}{1} & \multirow{2}{*}{1} \\
% & & & 不容易发胖 &  people who want to lose weight & & \\
\hline
\multirow{7}{*}{CVAE} 
& 信息 & information & 减肥的时候吃的热量低的食物 & low calorie food when losing weight & 1 & 1\\
\cline{2-7}
& \multirow{2}{*}{信息} & \multirow{2}{*}{information} & \multirow{2}{*}{减肥吃什么肉会胖} & what meat will lead to fat if I eat during & \multirow{2}{*}{0} & \multirow{2}{*}{1} \\
& & & & weight loss & & \\
\cline{2-7}
& \multirow{2}{*}{美容保健} & \multirow{2}{*}{beauty care} & \multirow{2}{*}{减肥期间吃什么主食最好} & what is the best staple food  & \multirow{2}{*}{1} & \multirow{2}{*}{1} \\
& & & & during weight loss & & \\
\cline{2-7}
& \multirow{2}{*}{电脑网络} & \multirow{2}{*}{Internet} & \multirow{2}{*}{一天中什么时候减肥最好} & when is the best time to lose  & \multirow{2}{*}{0} & \multirow{2}{*}{1} \\
& & & & weight in a day & & \\
\hline
\multirow{5}{*}{DCKG} 
& 美容保健 & beauty care & 减肥瑜伽视频 & weight loss yoga video & 1 & 1\\
\cline{2-7}
& 美容保健 & beauty care & 减肥小妙招 & weight loss trick & 1 & 1\\
\cline{2-7}
& 美容保健 & beauty care & 减肥能吃荔枝吗 & can I eat litchis during weight loss & 1 & 1\\
\cline{2-7}
& \multirow{2}{*}{信息} & \multirow{2}{*}{information} & \multirow{2}{*}{喝红豆薏米减肥吗} & is drinking red bean glutinous rice   & \multirow{2}{*}{1} & \multirow{2}{*}{1} \\
& & & & helpful to lose weight & & \\
\hline
\end{tabular}
\caption{Sample keywords generated by different models with domain accuracy (Acc.) and relevance (Rel.).}
\label{tab:sample_keywords}
\vspace{-6mm}
\end{table*}

\begin{table} [!htp]
\centering
\small
\begin{tabular}{|c|l|l|c|c|}
\hline
$\beta$ & \multicolumn{2}{c|}{Keyword} & Acc. & Rel. \\
\hline
\multirow{3}{*}{0.00} & Original & 减肥瘦身应该吃什么 & \multirow{3}{*}{0} & \multirow{3}{*}{1}\\
\cline{2-3}
& \multirow{2}{*}{Translated} & what should I eat to & & \\
& & lose weight and slim & & \\
\hline
\multirow{3}{*}{1.00} & Original & 减肥吃什么肉会胖 & \multirow{3}{*}{0} & \multirow{3}{*}{1}\\
\cline{2-3}
& \multirow{2}{*}{Translated} & what meat will lead to fat & & \\
& & if I eat during weight loss & & \\
\hline
\multirow{3}{*}{2.00} & Original & 减肥吃牛肉有什么好处 & \multirow{3}{*}{1} & \multirow{3}{*}{1}\\
\cline{2-3}
& \multirow{2}{*}{Translated} & the benefits of eating beef & & \\
& & during weight loss & & \\
\hline
\multirow{3}{*}{3.00} & Original & 减肥吃牛肉有什么好处 & \multirow{3}{*}{1} & \multirow{3}{*}{1}\\
\cline{2-3}
& \multirow{2}{*}{Translated} & the benefits of eating beef & & \\
& & during weight loss & & \\
\hline
\multirow{3}{*}{4.00} & Original & 减肥公司调查报告范文 & \multirow{3}{*}{1} & \multirow{3}{*}{0}\\
\cline{2-3}
& \multirow{2}{*}{Translated} & weight loss company survey & & \\
& & report essay & & \\
\hline
\multirow{3}{*}{5.00} & Original & 减肥公司调查调查信息 & \multirow{3}{*}{1} & \multirow{3}{*}{0}\\
\cline{2-3}
& \multirow{2}{*}{Translated} & weight loss company survey & & \\
& & survey information & & \\
\hline
\end{tabular}
\caption{Sample keywords conditioned on different $\beta$ with domain accuracy (Acc.) and relevance (Rel.).}
\label{tab:beta_keywords}
\vspace{-6mm}
\end{table}
\end{CJK}

Several sample keywords generated by different models are shown in Table \ref{tab:sample_keywords}. We can see that keywords generated by Seq2Seq have lower diversity than other models since it tends to generate keywords with the same prefix in similar meanings using beam search~\cite{li2015diversityMMI}, such as the 3rd and 4th keywords generated by Seq2Seq. 
Moreover, it can only predict one target domain category since it is a deterministic model, which cannot handle the task of generating different keywords from multi-domain categories. 

In comparison, CVAE and DCKG are able to predict multiple domain categories as well as generate diverse keywords according to these domain categories. However, in CVAE, the domain constraint is not addressed properly in some cases, as it takes a fixed way to combine the domain-specific information with the semantic information. For example, although the 6th generated keyword, ``what meat will lead to fat if I eat during weight loss'', is relevant to the source keyword, it is inconsistent with its domain category ``information'' since the domain-specific information is less considered than the semantic information in this case. By contrast, DCKG is able to choose a proper domain constraint factor to balance the domain-specific information and the semantic information for each instance. Hence, the generated keywords are both relevant to the source keyword and consistent with their domain categories.

In order to validate the influence of the domain constraint factor $\beta$ in the generated keywords, we presented several sample keywords generated by DCKG with different $\beta$ values in Table \ref{tab:beta_keywords}. Take the 6th keyword in Table \ref{tab:sample_keywords} as an example. Given the source keywords ``weight loss weight loss'' and its domain category ``beauty care'', our model predicts a target domain category ``information'', and then generates keywords with different $\beta$ values while other parameters are fixed. 

As shown in Table \ref{tab:sample_keywords}, when $\beta$ equals to 0.00, the target keyword belongs to the source domain category ``beauty care'' but violates the target domain category ``information'' since the keyword is generated regardless of the domain-specific information. 
With the increase of $\beta$, the domain accuracy improves since more domain-specific information is involved in the model, but the relevance decreases in some instances because the model faces with the compatibility of controlling semantic and domain-specific information. For example, the keywords generated when $\beta=2.00/3.00$, are both consistent to the target domain category and relevant to the source keyword. However, when $\beta$ increases to 4.00/5.00, the generated keywords are not relevant to the source keyword anymore, as a high $\beta$ value will harm the quality of generated keywords severely.

\section{CONCLUSION}
In this paper, we investigate the use of generative neural networks for ad keyword generation. We propose a Domain-Constrained Keyword Generator (DCKG) to generate diversified and domain-consistent keywords. 
A latent variable network is designed to model the diversity and a domain constraint network to satisfy the domain constraint in supervised learning. 
To further leverage the domain-specific information in keyword generation, we propose a reinforcement learning algorithm to adaptively utilize domain-specific information in keyword generation. Offline evaluation shows that the proposed model can generate keywords that are diverse, novel, relevant to the source keyword, and accordant with the domain constraint. Online evaluation shows that generative models can improve coverage (COV), click-through rate (CTR), and revenue per mille (RPM) substantially in sponsored search.

\section*{Acknowledgments}
This work was jointly supported by the National Science Foundation of China  (Grant No.61876096/61332007), and
the National Key R\&D Program of China (Grant No. 2018YFC0830200).

\bibliographystyle{ACM-Reference-Format}
\bibliography{www-bibliography}

\end{document}